\documentclass[letterpaper]{article} 
\usepackage{aaai25}  
\usepackage{times}  

\usepackage{helvet}  
\usepackage{courier}  
\usepackage[hyphens]{url}  
\usepackage{graphicx} 
\usepackage{natbib}  
\usepackage{caption} 
\usepackage{algorithm}
\usepackage{algorithmic}
\usepackage{amsmath, amsthm, siunitx} 
\usepackage{multirow}
\usepackage{booktabs}
\usepackage{amssymb}
\usepackage{csvsimple}
\newtheorem{theorem}{Theorem}

\usepackage{newfloat}
\usepackage{listings}
\DeclareCaptionStyle{ruled}{labelfont=normalfont,labelsep=colon,strut=off} 
\floatstyle{ruled}
\newfloat{listing}{tb}{lst}{}
\floatname{listing}{Listing}
\title{A New Dataset and Methodology for Malicious URL Classification}
\author {
    Ilan Schvartzman\textsuperscript{\rm 1},
    Roei Sarussi\textsuperscript{\rm 1},
    Maor Ashkenazi\textsuperscript{\rm 1,\rm 2},
    Ido Kringel\textsuperscript{\rm 1},
    Yaniv Tocker\textsuperscript{\rm 1},
   Tal Furman Shohet\textsuperscript{\rm 1}
}
\affiliations {
    \textsuperscript{\rm 1} Deep Instinct\\
    \textsuperscript{\rm 2} Department of Computer Science, Ben Gurion University of the Negev \\
    \{ilan.schvartzman, roeis, maor.ashkenazi, idok, yaniv.tocker, tal.furman\}@deepinstinct.com
}

\begin{document}

\maketitle

\begin{abstract}
Malicious URL (Uniform Resource Locator) classification is a pivotal aspect of Cybersecurity, offering defense against web-based threats. Despite deep learning's promise in this area, its advancement is hindered by two main challenges: the scarcity of comprehensive, open-source datasets and the limitations 
 of existing models, which either lack real-time capabilities or exhibit sub-optimal performance. In order to address these gaps, we introduce a novel, multi-class dataset for malicious URL classification, distinguishing between \textit{benign}, \textit{phishing}, and \textit{malicious} URLs, named \textbf{DeepURLBench}. The data has been rigorously cleansed and structured, providing a superior alternative to existing datasets. Notably, the multi-class approach enhances the performance of deep learning models, as compared to a standard binary classification approach. Additionally, we propose improvements to string-based URL classifiers, applying these enhancements to URLNet. Key among these is the integration of DNS-derived features, which enrich the model’s capabilities and lead to notable performance gains while preserving real-time runtime efficiency—achieving an effective balance for cybersecurity applications.

\end{abstract}

%
\begin{links}
     \link{Dataset}{https://github.com/deepinstinct-algo/DeepURLBench}
\end{links}

\section{Introduction}
\label{introduction}
Nowadays, internet browsing has become an essential aspect of our daily lives, encompassing activities such as social media engagement, business transactions, online shopping, and more.
This practice involves the usage of web addresses commonly referred to as URLs, functioning as the entry points to web pages and online resources. 
Each URL contains domain and top level domain and might include subdomains and file path, for example:
\begin{figure}[h!]
	\centering
	\includegraphics[width=\linewidth]{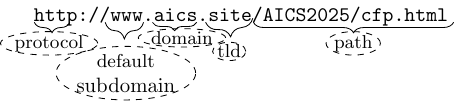}
	\label{fig:url_structure}
\end{figure}

Recent estimates indicate that there are over 1 billion web pages in existence today \citep{Siteefy}, approximately 1\% of which are categorized as malicious sources \citep{SecurityWeek}. As cybersecurity threats evolve, including tactics such as phishing\footnote{A website masquerading as a legitimate source, in order to steal information from users.}, malware\footnote{Malicious software.} distribution, and other online attacks, the development of effective URL classification methods becomes a top priority.

Machine learning methods have shown promise in tackling this challenge, yet their effectiveness is often limited by two key obstacles. First, the lack of extensive, openly accessible, and highly-curated datasets hinders the performance and generalizability of URL classification models. To address this gap, we introduce a novel dataset specifically designed to enhance malicious URL classification.

Second, evaluating the robustness of machine learning models over time remains difficult due to the rapid evolution of cybersecurity threats, which causes significant data distribution shifts. To address this issue, we propose a new benchmarking method to assess the robustness of URL classifiers over time. Our approach segments the test set by month, based on sample timestamps, to evaluate model performance across these time-based partitions. This analysis reveals model degradation over time and underscores the value of lightweight models, which can be retrained rapidly and cost-effectively to adapt to the shifting threat landscape.

Additionally, in recent years the field of Natural Language Processing (NLP) has achieved significant advancements, particularly through the introduction of the Transformer architecture \citet{vaswani2017attention}. Traditional models struggled with the inherent sequential nature of language and long-range dependencies between words. Unlike structured data, textual information follows less rigid patterns. Transformers, with their self-attention mechanisms, excel at capturing complex local and global dependencies within text and have been applied to malicious URL classification with promising results, as demonstrated by URLTran \citet{maneriker2021urltran}. However, the relatively large architecture of Transformers poses a challenge for real-time execution on consumer-grade CPUs, limiting their applicability in resource-constrained settings. Conversely, URLNet \citet{le2017urlnet}, a model primarily based on lightweight convolutional layers, achieves real-time performance but with lower accuracy as compared to URLTran.

Despite these advancements, deep learning models such as URLTran and URLNet have largely focused on extracting textual features from URLs, often overlooking essential non-linguistic features that have proven to be valuable in previous research. Notably, DNS data—which maps domain names to IP addresses—can provide critical server-related insights that aid in assessing the legitimacy of a URL. Additionally, lexical features, derived from expert knowledge, have historically contributed to distinguishing between malicious and benign URLs. Building on these insights, we propose a methodology that integrates these complementary features with modern deep learning techniques, combining textual and contextual data for a more comprehensive approach. By applying this methodology to URLNet, we significantly improve its classification accuracy, closing much of the gap with URLTran, while maintaining the efficiency required for real-time performance.

We summarize our contributions as follows:
\begin{itemize}
	\item We present a new meticulously curated dataset for malicious URL classification, named DeepURLBench. Our dataset is multi-class, categorizing URLs as either \textit{benign}, \textit{phishing}, or \textit{malware}.
    \item We establish a definition for \textit{real-time} URL classification models, based on web page statistics and user experience.
    \item We propose a new time-based benchmarking method, evaluating classifier robustness over time to address data distribution shifts.
    \item We develop an enhanced methodology that combines deep learning textual feature extraction with DNS and expert-crafted features, achieving improved classification accuracy while achieving real-time performance.
\end{itemize}

\section{Related Work}
\subsection{Existing Datasets}
\label{subsec:related_work_datasets}
In the domain of malicious URL classification, a major obstacle arises from the lack of comprehensive and publicly accessible datasets \citet{ya2019neuralas}. Datasets documented in previous studies suffer from various limitations. Some are not fully accessible to the general public \citet{le2017urlnet, tajaddodianfar2020texception, ma2009identifying}. Others incorporate URL shorteners \citet{wandhare2020phishing}, designed to transform long and complex URLs into shorter, more user-friendly links. For example, \textit{http://short.url/123456} could redirect the user to a long URL containing several subdomains and file paths. These have the potential to hide the true nature of any URL and need to be queried to reveal the underlying URL, needlessly complicating the classification task. Additionally, certain datasets predominantly feature URLs represented as IP addresses, or are of insufficient size for effective application in machine learning, making them inadequate for current research needs \citet{mahdavifar2021classifying}.

Creating a dataset in this field demands rigorous attention to avoid redundancy and ensure diversity \citet{tsai2022toward}. A prevalent issue is the recurrence of URL duplicates or the excessive representation of specific domains or subdomains. The approach taken by \citet{le2017urlnet} to mitigate this was to cap the presence of any single domain at 5\% of the training dataset. However, this strategy alone is insufficient to address the issue of generalization, as many URLs within these datasets originate from the same web pages with mere argument variations or share identical subdomains \citet{reynolds2022equivocal}, resulting in a relatively homogeneous dataset. This limits the dataset's ability to provide a comprehensive representation of the web landscape, hindering the development of robust malicious URL classification models. Another notable, publicly accessible, dataset is \textit{CIC-Bell-DNS 2021}, introduced in \citet{mahdavifar2021classifying}. This dataset exhibits an imbalance in domain representation, predominantly featuring a minimal assortment of domains for malicious URLs. The skewed distribution raises concerns about a machine learning model's potential to memorize specific malicious domains instead of learning to distinguish between benign and malicious URLs.

Furthermore, a critical aspect of dataset construction for malicious URL classification is the temporal separation between the training and test data. This ensures that the model is evaluated on its ability to predict future threats based on historical data. The dynamic nature of malicious URLs which are typically reported and subsequently removed or blacklisted, necessitates this temporal consideration \citet{han2016phisheye, drury2022dating,sheng2009empirical}. This issue is highlighted in Section~\ref{discussion:degradation}.


\subsection{Methods for Malicious URL Classification}
Malicious URL classification traditionally relies either on URL blocklists \citet{ma2009beyond}, or on classical machine learning approaches \citet{zhou2010high, lin2013malicious}. However, these methods can be circumvented by malicious actors through various means, namely URL obfuscation \citet{garera2007framework} or the use of Domain Generation Algorithms (DGA) \citet{antonakakis2012throw}. 
In addition to these vulnerabilities, blocklists also pose significant challenges in terms of memory efficiency, due to the exhaustive list of URLs, and lack of ability to generalize,
offering no protection against newly crafted URLs not already present in the list. These limitations have led to the exploration of other methods, including more advanced machine learning-based approaches.

Early works often applied manual feature extraction, where characteristics such as URL length, entropy of characters, and presence of specific tokens were used to train classifiers \citet{ma2009identifying,hwang2013classifying}. These features provided a fundamental understanding of malicious URLs, but required significant domain knowledge expertise and were limited in their ability to capture complex patterns.

Recent advancements have highlighted a shift towards leveraging deep learning for URL classification, extracting learned features from raw URL strings. In URLNet \citet{le2017urlnet}, a convolutional neural network (CNN) was designed to capture patterns within the URL, both at the character and at the word level. A similar method was proposed in Texception \citet{tajaddodianfar2020texception}, including a new architecture and benchmark, although neither were made public. Finally, URLTran \citet{maneriker2021urltran} proposed a Transformer network for URL classification. 
This enabled capturing complex global patterns within the data, improving classification performance. 
Among these approaches, only the initial one provides real-time capabilities, a conclusion supported by our analysis outlined in Section~\ref{background_runtime}.

\section{DeepURLBench}
\subsection{Data Sources}
\label{subsec:creation_criteria}
The URLs in our dataset are gathered from various sources, including publicly available deny-lists and allow-lists between the years 2020-2023. Our labeling criteria is based on \citet{virustotal}, a renowned online service that utilizes over $70$ cybersecurity vendors to scan and classify URLs for potential threats. This vast array of assessments ensures a comprehensive evaluation of each URL, making it a reliable source for classification.  

\subsection{Labels}
\label{subsec:labels_set}
We start by exploring different tags by vendor verdicts to reach the set of label categories. The most common tags by a large margin are 'clean' or 'unrated'; these are the default values any vendor yields (unless a different tag was found). Therefore, we focus our analysis on the non-safe tags to define our labeling criteria.

We define the \textit{potential non-safe URL's dataset} as all the URLs we gathered that had received at least one non-safe verdict from any vendor 

\begin{figure}[h!]
	\centering
	\includegraphics[width=\linewidth]{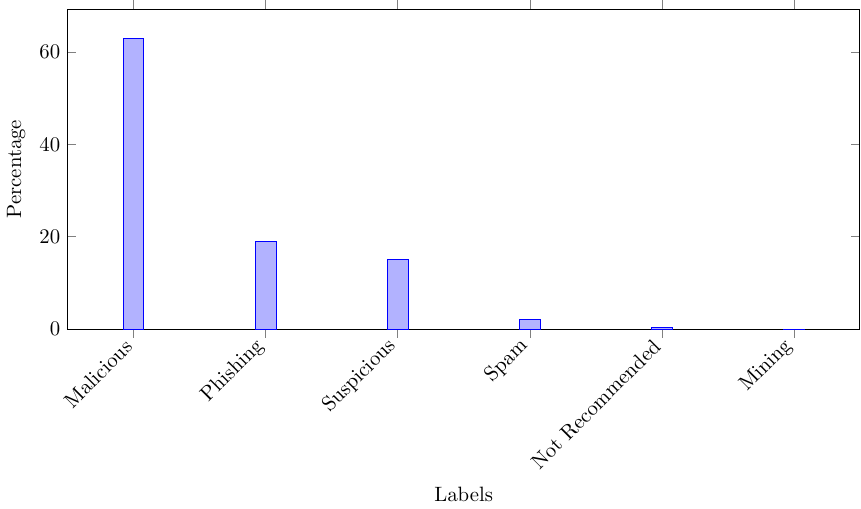}
	\caption{Histogram depicting the percentage of URLs detected as a non-safe tag by any number of vendors.}
	\label{fig:histogram_tags}
\end{figure}

Figure \ref{fig:histogram_tags} highlights that the non-safe significant  tags are 'Malicious' 'Phishing' and 'Suspicious'. We chose to discard the 'Suspicious' tag due to it potentially overlapping between the non-safe and safe tags. Finally we are left with our set of labels: \textit{benign} (corresponding to 'clean'), \textit{malware} and \textit{phishing}.

\subsection{Labeling Criteria}
A common method of labeling is majority voting \citet{raykar2010learning,donmez2009efficiently}, however, a quick data analysis shows that majority voting will always lead us to label a URL as a \textit{benign} site.

To label our dataset, we rely on a heuristic developed by experienced threat intelligence researchers. The process is guided by two key factors: detection-rate, ensuring the labeling criteria correctly assign the appropriate label to each URL, and coverage, ensuring the criteria are inclusive enough to encompass relevant URLs without being overly restrictive.

\subsubsection{Detection Rate}
\begin{figure}[h!]
\centering
	\includegraphics[width=\linewidth]{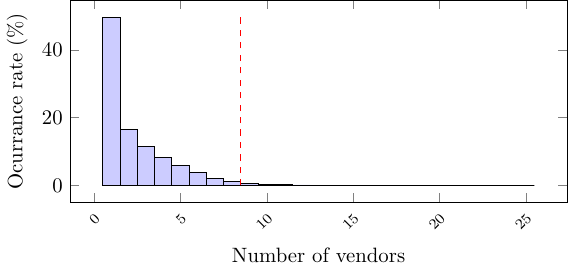}
    	\caption{Histogram of the occurrence rate of agreement between different vendors on the same verdicts out of all potentially non-safe URL's}
	\label{fig:vend_count_distribution}
\end{figure}

We create a calibration dataset consisting of high-confidence non-safe URLs. This dataset includes URLs that have 9 or more matching non-safe detections (top 1 percentile, Figure \ref{fig:vend_count_distribution}) from different vendors.

Using the calibration dataset we measure the detection rate of each vendor using the calibration dataset labels as a baseline.

\subsubsection{Coverage}
We count how many URLs are flagged as non-safe from each vendor out of the potential non-safe population.
    
    

Combining both coverage and detection rate, we yield a quality score for each vendor as a power mean of the coverage and detection rate. Vendors with over 90\% detection rate and 10\% coverage on the calibration dataset (high quality vendors) along side their quality score are given in Table \ref{fig:top_vendors_dr_on_calibration_set}.

\begin{table}
    \centering
    \begin{tabular}{l|c|c|c}%
    \bfseries Vendor & \bfseries \parbox[c][25pt][c]{45pt}{Detection \\ Rate}  & \bfseries Coverage & \bfseries Quality 
    \begin{filecontents*}{vendors_coverage_recall.csv}
Vendor,Recall,Coverage,Quality
Sophos,0.99,0.38,0.86
ESET,0.99,0.16,0.86
Fortinet,0.97,0.42,0.85
BitDefender,0.96,0.24,0.84
Kaspersky,0.96,0.20,0.83
G-Data,0.95,0.27,0.83
Lionic,0.95,0.18,0.83
CyRadar,0.94,0.27,0.82
alphaMountain.ai,0.93,0.14,0.81
Webroot,0.91,0.34,0.79
\end{filecontents*}
\csvreader[head to column names]{vendors_coverage_recall.csv}{}
    {\\\hline\Vendor & \Recall & \Coverage & \Quality}
    
    \end{tabular}
    \caption{High quality vendors metrics}
    \label{fig:top_vendors_dr_on_calibration_set}
\end{table}
\subsubsection{Labeling Criteria}
A URL is labeled as \textit{malware} or \textit{phishing} if it had unanimous non-safe verdicts from at least 2 high quality vendors and the overall detection quality (sum of the quality scores) crossed the quality threshold (2.5).
This stringent criterion enhances the reliability of our labels.
URLs not flagged as malicious by \textbf{any} high quality vendor are classified as \textit{benign} \citet{ya2019neuralas}.

\subsection{DNS Response}
A dimension often overlooked in existing datasets, is the inclusion of DNS response data.
DNS is a vital internet protocol that translates human-readable URLs into numerical IP addresses (e.g., 192.168.1.1).
It acts as a fundamental address book of the internet.
DNS queries occur naturally during web browsing. The Internet Service Provider (ISP) is responsible for answering those queries without requiring additional services or costs. 
Online public available data can associate each IP address with an Autonomous System Number (ASN), country, and Internet Service Provider (ISP),
all providing valuable information regarding web infrastructure. 
Thus, the DNS records provide a valuable source of information for URL classification. This is in contrast to other potential aggregated data sources which might incur additional costs. Therefore, for each URL we fetch the IP address from the DNS response and add it to the dataset.

\subsection{Preprocessing and Data Curation}
\label{subsec:Preprocessing}
We outline our five steps process for refining the dataset:

\textbf{URL Format Validation:} URLs either represented by IP addresses, or not beginning with \textit{http} or \textit{https}, were removed to ensure data consistency.


\textbf{Normalization of URLs:} URLs were transformed into their canonical form, removing queries and fragments for standardization \citet{berners2005uniform}.

\textbf{Top-Level Domain Validation:} To maintain data integrity, we removed URLs associated with top-level domains not officially recognized by the Internet Corporation for Assigned Names and Numbers (ICANN).

\textbf{DNS Responsiveness Check:} Unresponsive domain URLs were padded with a default value.

\textbf{Duplicate Entry Resolution:} Duplicates were resolved by retaining the URL with the earliest "first seen" timestamp.

\subsection{Temporal considerations}
\label{subsec:Dataset_dist}

Each URL in Virustotal has a first seen field which describes the first time this URL appeared in virus total. We use this field as an approximation for the time this URL appeared on the internet. 
\subsubsection{Reputation building time}
Security vendors do not tag non-safe URLs as they are created, most of the time. Hence, URLs with a timestamp of two months prior to collection are filtered out. This threshold is chosen to minimize the risk of including potentially harmful URLs that have not yet been identified as such. This selective process ensures that our dataset is comprised solely of URLs with clear and reliable labeling.

\subsubsection{Temporal separation}
\begin{figure}[h!]
	\centering
	\includegraphics[width=\linewidth]{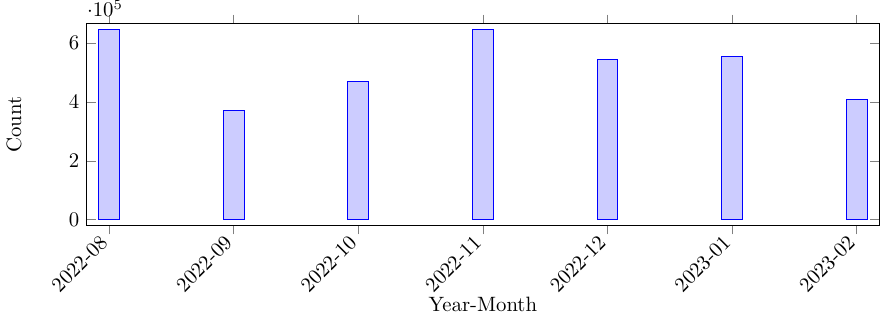}
	\caption{Histogram showing the number of URLs in the test set by their first appearance date in VirusTotal.}
	\label{fig:datetime_urls}
\end{figure}

A key aspect in cybersecurity is having a temporal separation between the training data and the test data for showing temporal generalization as suggested in \citet{boutaba2018comprehensive}. Therefore, for preventing any potential data leakage, URLs that first appeared on Virustotal before September 2022 were assigned to the training \& validation set while those that appeared after were assigned to the test set (Table \ref{tab:train_test_distribution_table}). Additionally, in the time-varying landscape of cybersecurity, threats are constantly evolving, it is imperative that any model trained on our dataset undergoes an evaluation process of performance degradation through time (view test set time distribution at Figure \ref{fig:datetime_urls})
\newline

\noindent To summarize, as outlined in Section ~\ref{subsec:related_work_datasets}, existing datasets for malicious URL classification are fraught with issues such as inadequate size, lack of diversity, temporal irrelevance, and the omission of valuable DNS response data, all of which impede the development of effective and generalizable malicious URL classification models. We address these issues in DeepURLBench.

\begin{table}[h!]

	\centering
	\begin{tabular}{cccc}
		\toprule
		\multirow{2}{*}{Split} & \multirow{2}{*}{Label} & \multirow{2}{*}{\shortstack{Number of \\ Samples}} & \multirow{2}{*}{\shortstack{Percentage \\ (\%)}} \\
		& & & \\
		\midrule
		\multirow{3}{*}{Total}  & Benign & 12,681,543 & 56.8 \\
		& Malware & 2,871,946 & 12.9 \\
		& Phishing & 6,764,542 & 30.3 \\
		\midrule
		\multirow{3}{*}{Train} & Benign & 10,539,529 & 57.8 \\
		& Malware & 1,989,811 & 10.9 \\
		& Phishing & 5,719,379 & 31.3 \\
		\midrule
		\multirow{3}{*}{Test} & Benign & 2,142,014 & 62.2 \\
		& Malware & 882,135 & 18.2 \\
		& Phishing & 1,045,163 & 19.6 \\
		\bottomrule
	\end{tabular}
    	\caption{Dataset distribution by class split into training (including validation) and test sets.}
	\label{tab:train_test_distribution_table}
\end{table}

\section{Holistic Design and Implementation}
\subsection{Real-Time Malicious URL Classification}
\label{background_runtime}
Runtime efficiency is a pivotal consideration in malicious URL classification, with direct impact on user experience in web browsing.
 ~\citet{arapakis2014impact} analyzed the perceived latency when browsing web pages, learning that most users do not notice added latency below \SI{0.5}{\second}.
Another issue to consider, is that web pages typically involve numerous URL requests. The returned resources might trigger additional URL requests, making the web browsing process sequential. 

\begin{figure*}[ht!]
	\centering
	\includegraphics[width=0.82\textwidth]{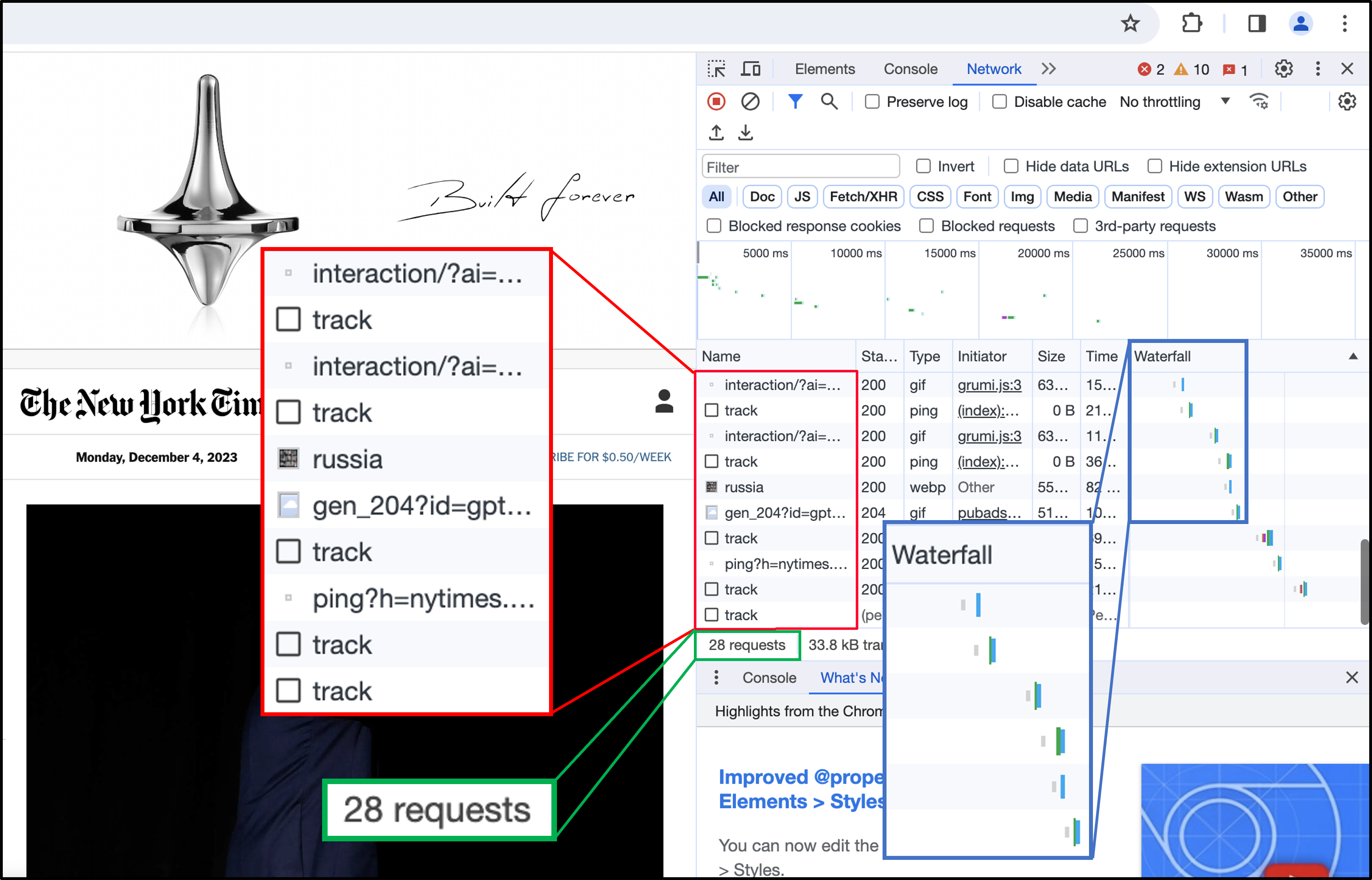}
	\caption{A browser screen with the developer tools panel. This shows a close-up view of requests initiated by the browser (marked in red), the total number of requests (marked in green) and the requests timeline (marked in blue). Note how a single web page request leads to additional URL requests, resulting in a substantial overall request count for the session.}
	\label{fig:browser_requests}
\end{figure*}

\begin{figure}[h!]
	\centering
	\includegraphics[width=\linewidth]{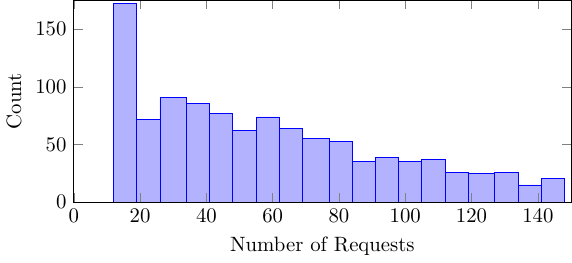}
	\caption{Histogram depicting the number of requests initiated when browsing the top 1500 websites worldwide, as ranked by \citet{alexa_top_million}.}
	\label{fig:histogram_requests}
\end{figure}

Figure~\ref{fig:browser_requests} demonstrates the requests sent by the browser to a specific web page behind the scenes. To give a sense of the amount of work the browser is performing, we analyze the number of requests initiated by browsing to the top 1500 domains worldwide, as ranked by \citet{alexa_top_million}. We find a median of $50$ and an average of $95$ unique requests per web page. The full histogram is depicted in Figure~\ref{fig:histogram_requests}. 
Combining these observations, we deduce the following:
\begin{theorem}
\label{theorem:realtime_classification}
For a malicious URL classification method to be classified as real-time, it must sequentially classify 50 URLs, in less than \SI{0.5}{\second} (or a single URL in less than \SI{10}{\milli\second} on average).
\end{theorem}

While trivial solutions such as block-lists offer efficiency in terms of latency, their inability to anticipate new or unknown malicious URLs is a notable shortcoming. In contrast, deep learning methods excel at generalizing from known data to unseen threats, though they often come with substantial computational costs that can hinder real-time performance. Our enhanced methodology, built on URLNet, achieves a balance between effectiveness and real-time performance, providing a robust solution for malicious URL classification.

\subsection{Architecture}
\begin{figure*}[ht!]
	\centering
	\includegraphics{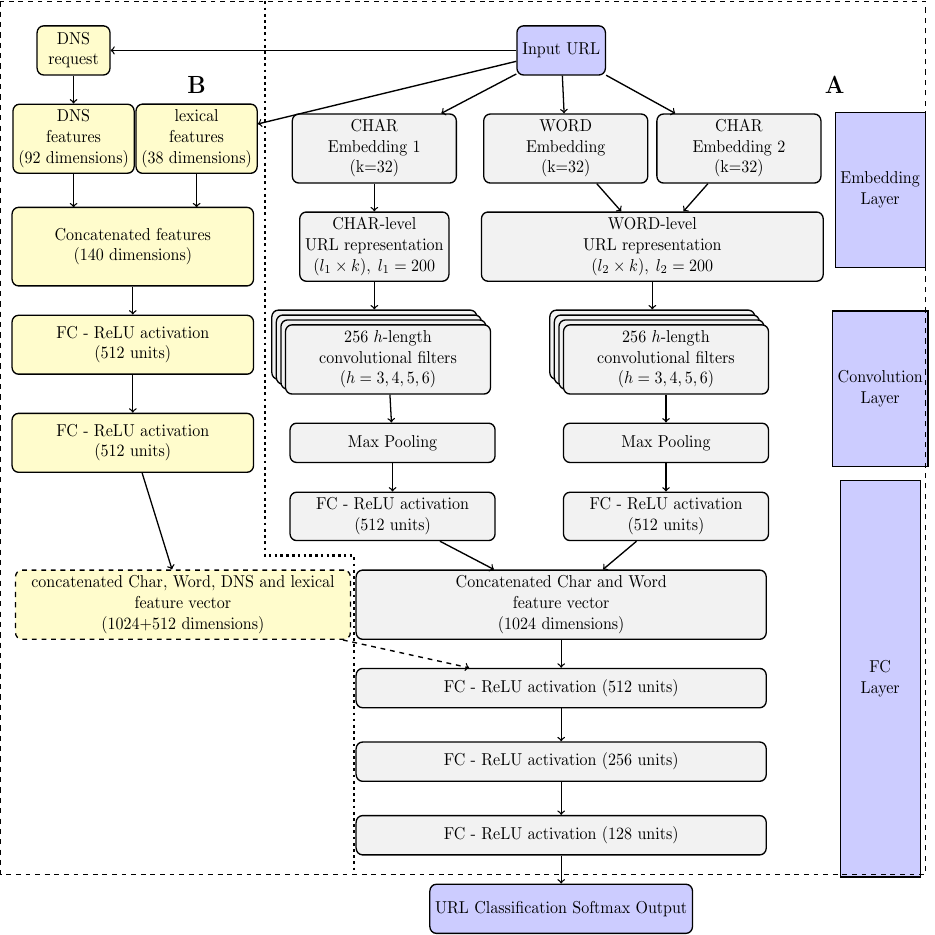}
	\caption{An overview of the suggested modifications for URLNet. An input URL (A) Original URLNet (B) URLNet\textsuperscript{+}}
	\label{fig:architecture}
\end{figure*}
Basically, our approach is inspired by \citet{iizuka2016let} that used global image features for improving grayscale coloration. We employ global URL features for improving URLNet. URLNet is composed of two branches, one processing the URL at the character level, and the other at the word level. Embeddings are learnt independently for characters and words. Before being fed to the network, the word encodings are enriched using character encoding information to create a combined word representation. Convolutional neural networks is a powerful tool for classification of data with strong locality features.  Furthermore, we propose two extensions based on DNS information and global lexical feature information. The original URLNet architecture is presented in Figure~\ref{fig:architecture}(A) and the modifications suggested (denoted as URLNet\textsuperscript{+}) in Figure~\ref{fig:architecture}(B).

\subsection{Incorporating Global Lexical Features}
Previous work has demonstrated that the inclusion of global lexical features can significantly improve the efficacy of malicious URL classification methods \citet{lin2020comparison, uto2020neural}. These features provide global contextual understanding within the initial stages of the network, enabling us to minimize the number of layers. 

We derive our global lexical features based on the approaches outlined in \citet{lin2020comparison, uto2020neural}. We exclude features of low importance based on their \textit{p-value}, using a threshold of 0.05.

Table~\ref{tab:lexical_features} presents the selected features. The global lexical features are processed using two fully connected layers, before being merged with the textual features extracted from the character CNN and word self-attention blocks.

\begin{table}[h!]
	\centering
		\begin{small}
			\begin{sc}
				\begin{tabular}{lll}
					\toprule
					URL part & feature & \\
					\midrule
					\midrule
					protocol    & 1 for https 0 for http \\
                    \midrule
					\multirow{5}{*}{domain} & number of `-` \\
					 & number of digits \\
					 & number of characters \\
					 & \multirow{2}{*}{\shortstack{entropy: $-\sum_{c \in C} P(c) \log_n P(c)$}} \\
                    & & \\
         		\midrule
					subdomains & number of subdomains \\
                    \midrule
					tld & one hot encoded \\
                    \midrule
					\multirow{8}{*}{Path} & number of `@` \\
					 & number of `\%` \\
					 & number of `*` \\
					 & number of `*` \\
					 & number of `\&` \\
					 & number of '(' \\
                      & number of ')' \\
				     & number of subdirectories\\
                    \midrule
                    \multirow{3}{*}{entire URL} & number of spaces \\
                    & \multirow{2}{*}{\shortstack{entropy: $-\sum_{c \in C} P(c) \log_n P(c)$}} \\
                    & \\

					\bottomrule
				\end{tabular}
			\end{sc}
		\end{small}
        \caption{Extracted global lexical features.}
 \label{tab:lexical_features}
\end{table}

\subsection{Incorporating DNS Features}
\label{subsec:dnsfeatures}

Using \citet{ip2asn} we associate each IP address with an Autonomous System Number (ASN), country, and Internet Service Provider (ISP). Based on their prevalence within DeepURLBench, we use the top 30 ASNs, countries, and ISPs, along with an \textit{other} class for each, to construct our DNS features. We use three boolean feature vectors, where each entry in a vector corresponds to a specific value of ASN, country or ISP. Since a DNS request can map a URL to more than one IP address, the binary feature vectors may have more than one active bit. We further add the number of mapped IP addresses, and Time To Live (TTL), as features. The DNS features are processed similarly to the global lexical features.

\begin{table*}[ht]

	\centering
		\begin{sc}
			\begin{tabular}{@{}ccccrrrrr@{}}

                \midrule
				Model&Lexical&Classes&DNS&AUC&Recall&Recall&Runtime&Num of\\
				& & &  &&@1\%&@0.1\%&($ms$) &Params  \\
				\midrule
				\midrule
				URLNet&x&binary&x&$0.983\pm 0.0007$&$78.0\pm0.004$&$52.1\pm0.006$&\phantom{0}0.332&7.9M \\
URLTran&x&binary&x&$0.987\pm0.0026$&$83.1\pm0.031$&$\pmb{57.4\pm0.03}$&41.751&109M \\
				\midrule
				\midrule
				URLNet\textsuperscript{+}&\checkmark&binary&x&$0.984\pm0.0004$&$79.0\pm0.004$&$53.0\pm0.004$&\phantom{0}0.525&8.08M\\
				URLNet\textsuperscript{+}&\checkmark&binary&\checkmark&$0.989\pm0.0004$&$83.2\pm0.004$&$55.6\pm0.006$& \phantom{0}0.542&8.13M \\
				URLNet&x&multi-class&x&$0.983\pm0.0004$&$78.2\pm0.006$&$52.8\pm0.005$&\phantom{0}0.334&7.8M \\
                URLNet\textsuperscript{+}&\checkmark&multi-class&x&$0.983\pm0.0004$&$79.0\pm0.005$&$53.2\pm0.003$&\phantom{0}0.525&8.13M \\
		URLNet\textsuperscript{+}&\checkmark&multi-class&\checkmark&$\pmb{0.988\pm0.0004}$&$\pmb{83.5\pm0.004}$&$56.4\pm0.007$&\phantom{0}0.542& 8.13M\\
				\bottomrule
			\end{tabular}
		\end{sc}
        	\caption{Classification results for URLNet with and without modifications, URLNet\textsuperscript{+}, URLTran on DeepURLBench. Real-time threshold requires runtime to be bellow \SI{10}{\milli\second}.}
 	\label{tab:results}
\end{table*}

\subsection{Multi-class Loss Function}
\label{subsec:multi-class_loss}
 While our primary goal is to \textbf{block all malicious URLs, be they malware or phishing}; it is worthwhile to explore whether a multi-class approach could yield better performance than standard binary classification approach. Rather than using a \textit{softmax} function for multiple class probabilities, we use a two step solution, applying two binary cross-entropy loss functions. A similar methodology was previously explored for object detection tasks \citet{girshick2015fast}. The first binary loss differentiates between \textit{benign} and \textit{malicious}, while the second differentiates between the specific type of maliciousness - either \textit{phishing} or \textit{malware}. The additional granularity offered by the multi-class approach enhances the capabilities of our model, as illustrated by the results in Section~\ref{exp}. This methodology is also useful for cases where the dataset contains malicious samples, without a clear differentiation between \textit{phishing} and \textit{malware}.


\section{Experimental Results}
\label{exp}
We train URLNet and URLTran on DeepURLBench and evaluate the results using the Area Under Curve (AUC) and Recall @ $1\%$ False Positive Rate (FPR), similarly to \citet{le2017urlnet, maneriker2021urltran}. The metrics reported in Table~\ref{tab:results} represent the average performance over five models trained with different random seeds, with standard deviations denoted using the $\pm$ symbol. This provides a robust assessment of performance consistency across multiple runs. 

\begin{figure}[h]
	\centering
	\includegraphics[width=\linewidth]{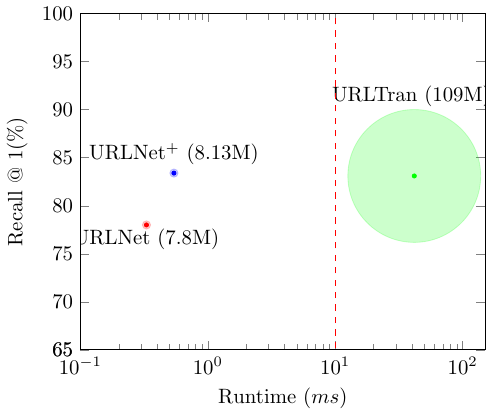}
\caption{Runtime comparison of the evaluated models. Circle radius is proportional to the number of parameters, indicated in parenthesis. The dashed  line stands for the real-time threshold. Note that the runtime axis is log-scaled.}
	\label{fig:runtime}
\end{figure}

We evaluate model runtime after compiling to ONNX \citet{bai2019}. The runtime measurements were conducted on an Intel(R) Core(TM) i7-10700 CPU @ 2.90GHz. The results in Table~\ref{tab:results} and Figure~\ref{fig:runtime} demonstrate that URLNet's performance can be significantly enhanced without a substantial increase in runtime.

The results on DeepURLBench differ from previous works due to the temporal diversity of our dataset, which spans three years and reflects evolving URL patterns, presenting a more realistic yet challenging classification problem.


\section{Discussion}
\subsection{Model Degradation Through Time}
\label{discussion:degradation}
\begin{figure}[h]
	\centering
	\includegraphics[width=\linewidth]{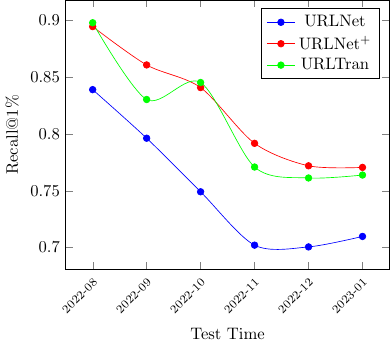}
	\caption{Malicious URL classification model's degradation through time.}
	\label{fig:degradation}
\end{figure}

A common characteristic in cybersecurity is the rapid evolution of threats, comprising a severe data distribution shift challenge. To analyze this effect in malicious URL classification, we partition the test set, and evaluate the models on 1 month segments of the test data, based on sample's timestamps. The results, presented in Figure~\ref{fig:degradation}, reveal a distinct pattern of model degradation through time. In such a dynamic landscape, the importance of a lightweight model becomes clear. Models with fewer parameters can be retrained rapidly and at lower cost, enabling a more agile response to the evolving threat environment. Note that while the performance of all models degrade, URLNet\textsuperscript{+} consistently outperforms URLNet and at times even manages to surpass URLTran, a large Transformer model.

\subsection{Limitations and Challenges}
DNS data allows for notable improvement in efficacy. Nevertheless, DNS querying can be influenced by load balancing infrastructure, leading to varied responses when requests originate from different ASNs or are made at different times. While techniques exist to overcome these challenges, such as using specialized services to retrieve all possible responses, we did not employ such methods in this research. Instead, we relied on standard DNS queries as they would normally be performed. Therefore, while this limitation may exist in theory, it does not directly impact the approach taken in this paper. Another challenge lies in the potential for models using DNS data to inadvertently develop biases towards certain geographical regions. Countries with a higher prevalence of malware or phishing attempts might be unfairly over-represented in the classification process. This highlights the need for careful consideration and continuous monitoring to maintain ethical cybersecurity practices.

\section{Future Work}
Several avenues for future work emerge from this study. Firstly, based on the observed degradation patterns, a key direction for further research is the application of robustness techniques and the investigation of features that can improve the model's stability and performance over time. In addition, it would be valuable to explore features that are more closely associated with benign websites, such as page ranking and site interlinking patterns. Lastly, we aim to investigate the potential integration of online learning approaches. This would allow the model to adapt continuously to new threats, ensuring sustained high performance as the landscape of malicious URLs evolves.

\section{Conclusions}
In this paper, we introduce a large-scale, multi-class URL dataset for malicious URL detection, which, to our knowledge, is the first of its kind. We provide a detailed explanation of the labeling methodology, ensuring transparency and reproducibility. We added an additional dimension of data - the DNS response, for each URL dataset. Additionally, we incorporate a temporal separation between the training and testing sets to evaluate model performance over time. By including the timestamp of the URL's first appearance, we facilitate a series of temporal test splits, allowing for an analysis of model degradation as the dataset evolves. We added a definition for real-time URL detection based on user experience and browsing request statistics. These temporal considerations, namely, real-time and degradation rate are providing a new evaluating paradigm for URL detection systems. In terms of modeling, we combined the two most common paradigms, textual and global.  Our approach leverages URLNet, a convolutional neural network (CNN)-based model, which focuses on local features extracted from the URL itself, and is further enhanced by global features. These global features comprise both traditional lexical characteristics and DNS-related data, providing a more comprehensive representation of the URLs for improved classification accuracy, and lower degradation rate while maintaining real-time classification.   

\bibliography{aaai25}

\end{document}